\documentclass[10pt,twocolumn,letterpaper]{article}

\usepackage{iccv}
\usepackage{times}
\usepackage{epsfig}
\usepackage{graphicx}
\usepackage{amsmath}
\usepackage{amssymb}
\usepackage{amsthm}

\usepackage{dblfloatfix}
\usepackage{graphicx}
\usepackage{amsmath}
\usepackage{amssymb}
\usepackage{booktabs}
\usepackage{multirow}
\usepackage{multicol}
\usepackage{adjustbox}
\usepackage{subcaption}
\usepackage[ruled]{algorithm2e}
\usepackage[noend]{algpseudocode}
\usepackage{breqn}

\SetKwInput{KwInput}{Input}                
\SetKwInput{KwOutput}{Output}              

\newcount\Comments  
\usepackage{color}
\newcommand{\kibitz}[2]{\ifnum\Comments=1\textcolor{#1}{#2}\fi}
\newcommand{\ming}[1]{\kibitz{red} {[Ming:~#1]}}

\usepackage[breaklinks=true,bookmarks=false]{hyperref}

\iccvfinalcopy 

\def\iccvPaperID{5276} 

\ificcvfinal\fi

\begin{document}

\title{Better May Not Be Fairer: A Study on Subgroup Discrepancy \\ in  Image Classification}

\author{Ming-Chang Chiu\\
University of Southern California\\
Los Angeles, CA\\
{\tt\small mingchac@usc.edu}
\and
Pin-Yu Chen\\
IBM Research\\
Yorktown Heights, NY\\
{\tt\small pin-yu.chen@ibm.com}
\and
Xuezhe Ma\\
University of Southern California\\
Los Angeles, CA\\
{\tt\small xuezhema@isi.edu}
}

\maketitle
\ificcvfinal\thispagestyle{empty}\fi


\begin{abstract}
   In this paper, we provide 20,000 non-trivial human annotations on popular datasets as a first step to bridge gap to studying how natural semantic spurious features affect image classification, as prior works often study datasets mixing low-level features due to limitations in accessing realistic datasets. We investigate how natural background colors play a role as spurious features by annotating the test sets of CIFAR10 and CIFAR100 into subgroups based on the background color of each image. We name our datasets \textbf{CIFAR10-B} and \textbf{CIFAR100-B}\footnote{Dataset is released at https://github.com/charismaticchiu/Better-May-Not-Be-Fairer-A-Study-Study-on-Subgroup-Discrepancy-in-Image-Classification} and integrate them with CIFAR-Cs.
   
   \ming{re-written}We find that overall human-level accuracy does not guarantee consistent subgroup performances, and the phenomenon remains even on models pre-trained on ImageNet or after data augmentation (DA). To alleviate this issue, we propose \textbf{FlowAug}, a \emph{semantic} DA that leverages decoupled semantic representations captured by a pre-trained generative flow. Experimental results show that FlowAug achieves more consistent subgroup results than other types of DA methods on CIFAR10/100 and on CIFAR10/100-C. Additionally, it shows better generalization performance.
   
   Furthermore, we propose a generic metric, \emph{MacroStd}, for studying model robustness to spurious correlations, where we take a macro average on the weighted standard deviations across different classes. We show \textit{MacroStd} being more predictive of better performances; per our metric, FlowAug demonstrates improvements on subgroup discrepancy. Although this metric is proposed to study our curated datasets, it applies to all datasets that have subgroups or subclasses.  Lastly, we also show superior out-of-distribution results on CIFAR10.1. 
\end{abstract}

\vspace{-5mm}
\section{Introduction}

\begin{figure*}[t]
     \centering
     \includegraphics[width=\textwidth]{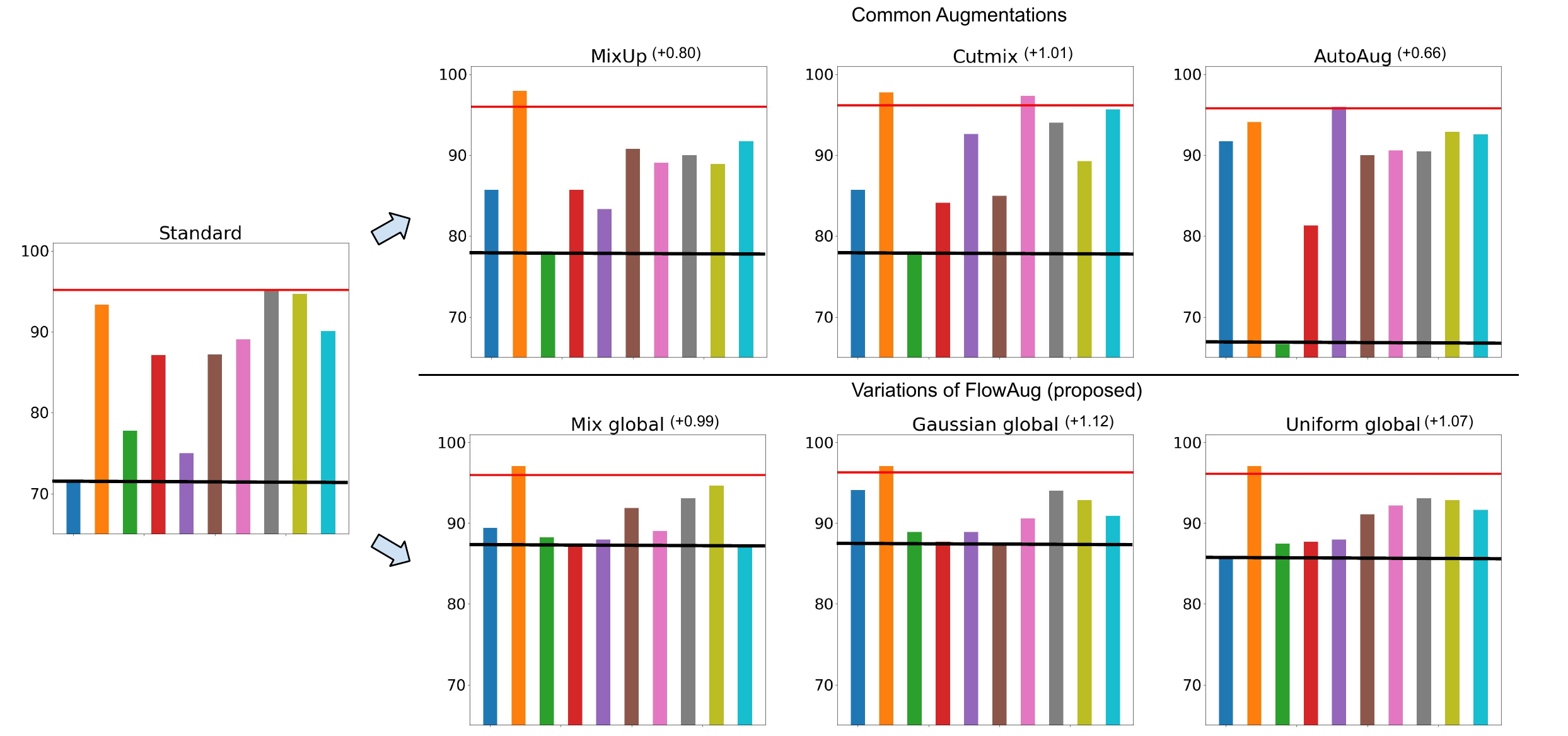}
        \vspace{-6mm}
        \caption{\textbf{FlowAug reduces subgroup discrepancy.} 
        \textbf{CIFAR10-B} enables us to observe the worst test time subgroup accuracy in each class. Standard ERM shows \textit{subgroup discrepancy}, uneven subgroup performances across all classes, and a huge gap between total accuracy (\textit{red line}) and the worst subgroup accuracy (\textit{black line}). This issue persists even after common DAs are used (\textit{top}). Our proposed \textbf{FlowAug} mitigates this issue (\textit{bottom}) and also reports improved overall performance.}
        \label{fig:teaser}
\end{figure*}

Deep neural networks (DNNs, e.g., \cite{lecun2015deep,He2016DeepRL}), properly trained via empirical risk minimization (ERM), have been demonstrated to significantly improve benchmark performances in a wide range of application domains.
However, minimizing empirical risk over finite or biased datasets often results in models latching on to \emph{spurious correlations} that do not show a robust relationship between the input data and output labels.
Moreover, benchmark evaluations based solely on average accuracy may overlook these critical issues. For instance, Fig.~\ref{fig:teaser} shows that on CIFAR10, even though a standard ERM model reaches human-level test accuracy (\textit{red line}), if we dive deeper into each class and compute their respective worst test accuracy stratified by background colors, they are inconsistent across the ten classes and the degradation from total accuracy is huge (\textit{black line}) for some. Such inconsistency and discrepancy have huge real-world implications, suggesting DNN models may make biased decisions against or in favor of specific spurious factors, such as certain background colors. 

\begin{figure*}[t]
     \centering
     \includegraphics[width=0.9\textwidth]{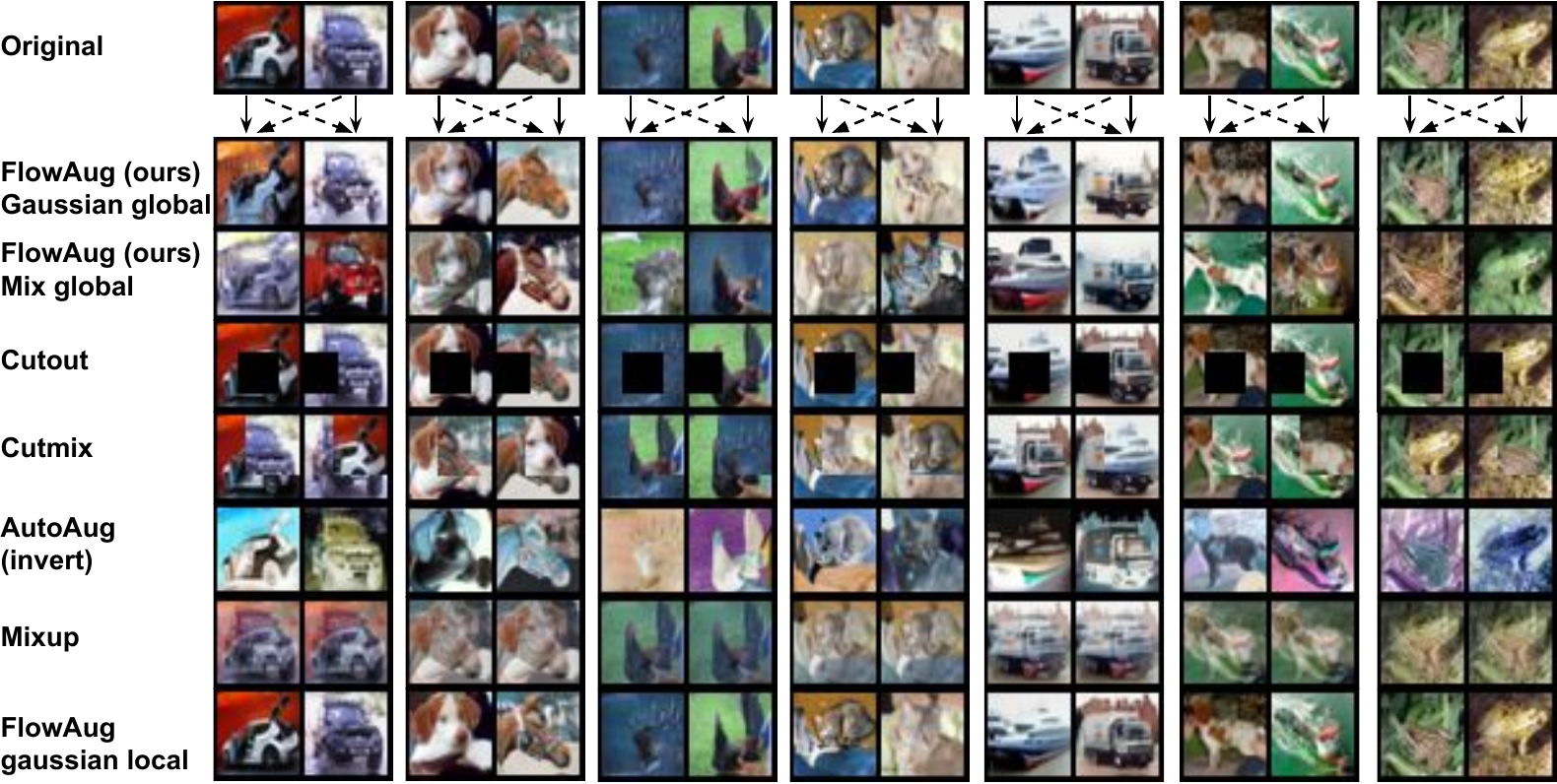}
        \vspace{-2mm}
        \caption{\textbf{Examples of different augmentation methods. } Row 2 \& 3 are generated by our methods.}
        \label{fig:augmentations}
        \vspace{-2mm}
\end{figure*}

Researchers have been working in different directions to understand the effect of spurious correlations, including model over-parameterization~\cite{sagawa2020investigation}, causality~\cite{Arjovsky2019InvariantRM} and information theory~\cite{lovering2020predicting,zhou2021examining}. 
Various techniques have emerged over the years to address this challenge, among which DA~\cite{shorten2019survey} has stood out for its simplicity and effectiveness. 
DA shows better generalization results in various machine learning tasks than other approaches~\cite{Zhang2018mixupBE, Yun2019CutMixRS, wei-zou-2019-eda, Guo2019AugmentingDW, Wang2021MultiFormatCL, Shen2020ASB}. 
These augmentation methods, however, are often based on heuristic and coarse image processing techniques such as flipping, rotating, blurring, 
or manipulating images by mixing attributes from other inputs \cite{Zhang2018mixupBE, Yun2019CutMixRS, Devries2017ImprovedRO, Hendrycks2020AugMixAS} (Fig.~\ref{fig:augmentations}); therefore, they can only address limited aspects of spurious correlations, for which we will show an example in \S~\ref{sec:case}. To address this limitation, instead of mixing low-level features, we seek to augment the training set by learning \emph{semantic} deep representations and then using them to generate new images. 

\begin{figure}[t]
     \begin{subfigure}[b]{0.1035\textwidth}
         \centering
         \includegraphics[width=\textwidth]{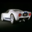}
         \caption{black}
     \end{subfigure}\hfill
     \begin{subfigure}[b]{0.1035\textwidth}
         \centering
         \includegraphics[width=\textwidth]{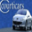}
         \caption{blue}
     \end{subfigure}\hfill
     \begin{subfigure}[b]{0.1035\textwidth}
         \centering
         \includegraphics[width=\textwidth]{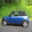}
         \caption{green}
     \end{subfigure}\hfill
     \begin{subfigure}[b]{0.1035\textwidth}
         \centering
         \includegraphics[width=\textwidth]{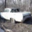}
         \caption{gray}
     \end{subfigure}
     
     \begin{subfigure}[b]{0.1035\textwidth}
         \centering
         \includegraphics[width=\textwidth]{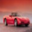}
         \caption{red}
     \end{subfigure}\hfill
     \begin{subfigure}[b]{0.1035\textwidth}
         \centering
         \includegraphics[width=\textwidth]{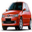}
         \caption{white}
     \end{subfigure}\hfill
     \begin{subfigure}[b]{0.1035\textwidth}
         \centering
         \includegraphics[width=\textwidth]{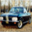}
         \caption{brown}
     \end{subfigure}\hfill
     \begin{subfigure}[b]{0.1035\textwidth}
         \centering
         \includegraphics[width=\textwidth]{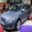}
         \caption{others}
     \end{subfigure}
     \vspace{-2mm}
     \caption{\textbf{Examples of CIFAR10-B (Car).} We label seven common background colors for CIFAR10 and CIFAR100. Difficult examples are categorized as ``others''.}
     \label{fig:subgroups_car}
     \vspace{-5mm}
\end{figure}

In this paper, as the very \emph{first} step towards comprehensive evaluation of subgroup performance against \emph{semantically meaningful and realistic spurious correlations} in image classification, we conduct a case study experiment to investigate background colors as spurious features (\S\ref{sec:case}), for their commonality in image classification and immediate implications for trustworthiness \cite{10.1145/2939672.2939778}. To directly quantify the results, we annotated the test data of CIFAR10 and CIFAR100 into subgroups based on natural image background colors (see Fig.~\ref{fig:subgroups_car}), yielding \textbf{CIFAR10-B}ackground and \textbf{CIFAR100-B}ackground. \ming{new argument}To the best of our knowledge, our datasets are \emph{two of the only human-annotated} benchmark datasets with a \emph{natural semantic bias}. \ming{completely new}We argue that the background color bias should be a \emph{necessary spurious correlation} for future studies on robustness to benchmark on and so our work can facilitate future works to benchmark their capabilities on reducing learning spurious factors. 
Equipped with our datasets, we can investigate the reliance on background color of deep neural models in a multi-class multi-subgroup setup. 

\begin{figure*}
  \centering
     \includegraphics[width=1\textwidth]{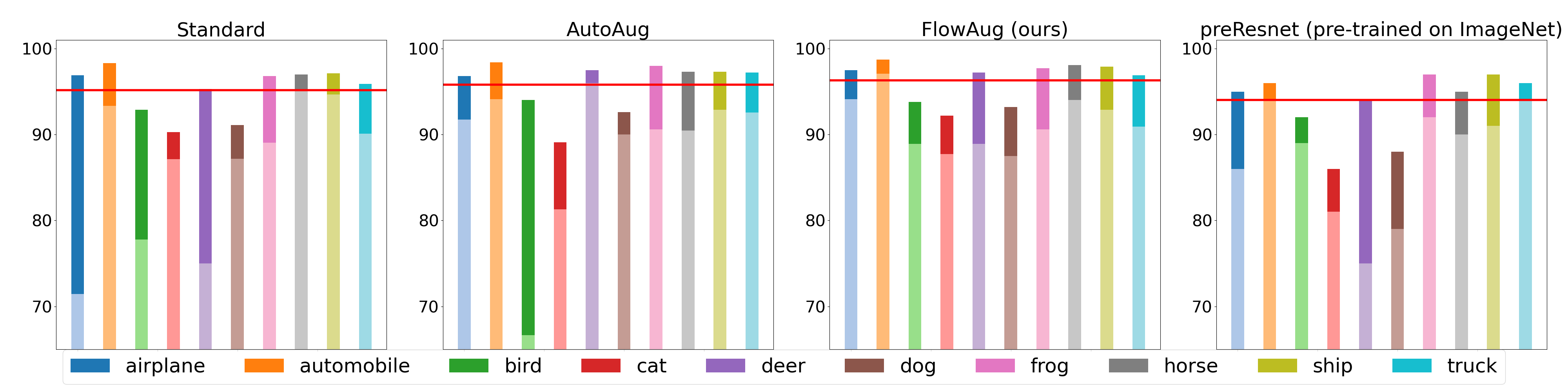}
        \vspace{-6mm}
        \caption{\textbf{Gaps between class accuracies (dark bars) and their worst subgroup accuracies (light bars).} Although a standard CNN model can reach human-level accuracy (\textit{red line}), we find that the subgroup performances can be surprisingly low. Even after data augmentation (\textit{mid-left}) or fine-tuned from ImageNet (\textit{right}), the same phenomenon remains. FlowAug \textit{(mid-right) shows more consistent results and mitigates the performance gaps (dark bars) the most}.\ming{added pretrained resnet18}} 
    \label{fig:case_study}
    \vspace{-2mm}
\end{figure*}

We reveal that even though standard DNNs have achieved human-level accuracy in image classification tasks, the performances fluctuate across different subgroups. This phenomenon demonstrates the reliance on background colors as spurious features.
Moreover, applying some popular DA methods or pre-training on larger dataset such as ImageNet do not prevent the models from producing uneven accuracies across subgroups, as shown in Fig.~\ref{fig:teaser}~\&~\ref{fig:case_study}, which further shows that low-level feature manipulations or brute-force pre-training are not sufficient to address spurious correlations and better methods are needed. To quantify our observations, we propose \emph{MacroStd}, a metric to quantify subgroup performance discrepancy and imply the reliance on spurious correlations (\S~\ref{sec:metric}). 


To enable semantic data augmentations and address the issue of uneven accuracies, we propose \textbf{FlowAug}, a novel DA method which is capable of manipulating images semantically via decoupled representations learned from invertible generative flows~\cite{decoupling2021} (\S\ref{sec:flowaug}). Concretely, our deep generative augmentation approach incorporates a novel flow-based generative model that encourages disentanglement of local and global representations from images, which arguably correspond to the image  ``style'' and ``content'' \cite{Gatys2015ANA, CycleGAN2017}, respectively. 
By operating on the global representation that is isolated with the image class label, FlowAug semantically creates new images for DA.

More consistent performance across subgroups demonstrates the effectiveness of FlowAug. Also, we integrate our CIFAR-Bs with CIFAR-Cs~\cite{hendrycks2019robustness} for broader out-of-distribution (OOD) evaluations and observe similar consistent subgroup performances\ming{rewritten}. Furthermore, though not our main foci, we also find that superior experimental results on various in-distribution (ID) and OOD benchmarks, including CIFAR10, CIFAR100~\cite{Krizhevsky2009LearningML}, CIFAR10.1\cite{recht2018cifar10.1, torralba2008tinyimages} bolster our belief that low-level manipulations or brute-force pre-training are not sufficient.

    
  
 To summarize, our contributions are four-fold, 
 \begin{itemize}
    \vspace{-1mm}
    \item We curate 20,000 human-annotated labels and two \textbf{CIFAR-B} datasets that reveal the \emph{subgroup discrepancy} phenomenon and allow us to (1) study semantically meaningful and realistic spurious correlations in a multi-class multi-subgroup setup, and (2) integrate with CIFAR-C for OOD evaluations. \ming{rewritten}
    \vspace{-1mm}
    \item We propose \textbf{FlowAug}, a novel augmentation method that leverages ``expert knowledge'' of the deep generative model to change semantic attributes of images and empirically shown to reduce subgroup discrepancy.  
    \vspace{-1mm}
    \item We propose a generic metric that captures the subgroup discrepancy phenomenon of ERM and common DA methods and measures the sensitivity of model performances to spurious correlations, and demonstrate FlowAug's effectiveness in this regard.  \ming{rewritten} 
    \vspace{-1mm}
    \item As an additional benefit, we conduct experiments on CIFAR10/100 and CIFAR10.1 and show FlowAug can further provide superior performances on ID and OOD datasets. 

 \end{itemize}
\begin{figure}[t]
     \begin{subfigure}[b]{0.23\textwidth}
             \centering \includegraphics[width=0.45\textwidth]{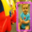}\enspace
             \includegraphics[width=0.45\textwidth]{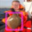}
         \caption{Principle 1}
     \end{subfigure}\hfill
     \begin{subfigure}[b]{0.23\textwidth}
             \centering
             \includegraphics[width=0.45\textwidth]{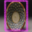}\enspace
             \includegraphics[width=0.45\textwidth]{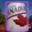}
         \caption{Principle 2}
     \end{subfigure}
     
     \begin{subfigure}[b]{0.23\textwidth}
             \centering
             \includegraphics[width=0.45\textwidth]{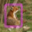}\enspace
             \includegraphics[width=0.45\textwidth]{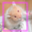}
         \caption{Principle 3}
     \end{subfigure}\hfill
     \begin{subfigure}[b]{0.23\textwidth}
             \centering
             \includegraphics[width=0.45\textwidth]{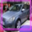}\enspace
             \includegraphics[width=0.45\textwidth]{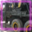}
         \caption{Principle 4}
     \end{subfigure}
     \vspace{-2mm}
     \caption{\textbf{Examples of the four main principles of our labeling philosophy.} See \S~\ref{sec:philosophy} for philosophy descriptions.}
     \vspace{-5mm}
     \label{fig:label_phil}
\end{figure}

\vspace{-1mm}
\section{A Motivating Example of Subgroup \\ Discrepancy}\label{sec:case}

We investigate background color as the spurious correlation with our CIFAR10-B (\S~\ref{sec:philosophy}) by first training a standard Resnet18 for 250 epochs with weight decay $5\times 10^{-4}$, initial learning rate 0.1 and learning rate decay at [100, 150] epochs by a factor of 0.1. We observe significant performance degradation in the subgroups of some classes, for example class ``airplane," ``bird" and ``deer" (Fig.~\ref{fig:case_study} (left)). Moreover, after applying DAs such as AutoAug \cite{Cubuk2018AutoAugmentLA}, the same phenomenon remains, for instance observe the ``bird" class in the mid-left plot in Fig.~\ref{fig:case_study}. \ming{new exp on preTrained Resnet}More surprisingly, even after we fine-tune Resnet18 pre-trained on ImageNet (preResnet) with similar protocol to \cite{BiT}, the degradation continues to exist (Fig.~\ref{fig:case_study} (right)).

In summary, though a Resnet model with or without popular DAs achieve more than 90\% class accuracies, their respective background subgroup performances can be surprisingly low. This phenomenon, which we call ``subgroup discrepancy" or ``in-class variability," shows that background colors play a role in the performance of a standard DNN model and constitute spurious correlations; otherwise, the performances should be relatively consistent. This triggers our interest in mitigating the performance variability in subgroups, i.e. the reliance on background attributes. And we show in Fig.~\ref{fig:teaser}~\&~Fig.~\ref{fig:case_study} that using FlowAug achieves more consistent subgroup results.

Furthermore, fine-tuning model pre-trained on larger benchmark such as ImageNet does not reduce \emph{subgroup discrepancy} even on dataset like CIFAR, so we reasonably conjecture that this phenomenon will exist in other datasets.

\vspace{-1mm}
\section{Methods}\label{sec:flowaug}

In principle, DA takes the form of a particular set of transformation functions $T$ where each $t\sim T$ transforms an input $x$ in a particular fashion. Moreover, an expert may have the knowledge to design label-preserving transformations $T$ in a way that $t(x)$'s leave the label unchanged. 

After the transformations, the dataset $\mathcal{D}$ will be augmented to $\{(x_i^{1:K},y_i)\}_{i=1}^{n}$, where $K$ is the number of times $x_i$ is transformed. From a frequentist point of view, we can apply any MLE algorithm to the augmented dataset, and the hope is that the learned model can better estimate the true model since we have more data. 


In this section, we discuss our generative flow model, present our datasets CIFAR10-B and CIFAR100-B for studying spurious correlation, and detail our augmentation algorithms. Lastly, we introduce two metrics to quantify the effect of spurious correlation.

\vspace{-1mm}
\subsection{Decoupling representations with Flow-based Generative Models}\label{sec:flow_property}



Prior work has shown that embedding a invertible normalizing flow model as a decoder in a variational autoencoder (VAE) can decouple global ($z$) and local ($\nu$) representations of images in an unsupervised fashion \cite{decoupling2021}, and can \emph{switch} the decoupled representations of different images to alter their semantic attributes (see Appendix). We presume the global information corresponds to the \textit{style} of the image and local leans toward the \textit{content} in the neural style transfer literature \cite{Gatys2015ANA, CycleGAN2017}. In this work, we apply the flow model $\mathcal{F}$ 
to encode images into global and local representations and also decode them back to image space like VAEs, 
\vspace{-2mm}
\begin{equation}\label{eq:encoder_decoder}
    z,\nu \leftarrow \mathcal{F}_{enc} (x);~~x' \leftarrow \mathcal{F}_{dec} (z,\nu) 
\end{equation}
where $z \sim \mathcal{N}(\mu(x)), \sigma(x))$, $\mu(x)$ and $\sigma(x)$ are neural networks learned from the data, and $\nu \sim \mathcal{N}(0, I)$. $z$ is a $d_z$-dimensional vector where $d_z$ is the dimension of the latent space and the size of $\nu$ is the same as the input image $x$.

We further hypothesize that $z$ includes information on colors or more, which are spurious to the ground truth, and $\nu$ bears information about the shape, object, etc., which are more indicative of the labels. In \S~\ref{sec:ablation_loc_v_global}, we will do an ablation study to attest this hypothesis.

\vspace{-1mm}
\subsection{Datasets quantifying spurious background correlations }\label{sec:philosophy}
We curate CIFAR-10-B \& CIFAR-100-B to identify and study spurious information in images, and we choose to label the major background colors of CIFAR10 and CIFAR100 validation sets. By learning the subgroup performances, we can measure the sensitivity of a model to different spuriously correlated colors. As shown in Fig.~\ref{fig:subgroups_car}, we manually label the background colors of CIFAR10 and CIFAR100, and split them into eight separate groups. We understand people have different criteria toward determining the background color; therefore, we provide our four main labeling principles as follows,

\vspace{-1mm}
\begin{enumerate}
    \item We label the color that has the most coverage around the object. In Fig.~\ref{fig:label_phil} (a), one may argue the red patch or blue ocean has taken up most of the image in the background, but the ``baby" is surrounded completely by the green area, and the ``flatfish" is in the red area.
    \vspace{-2mm}
    \item When two colors take almost the same coverage other than the object, we choose the color that appears further away. In Fig.~\ref{fig:label_phil} (b), black is farther away from the ``bowl", and so is the blue sky for the ``can".
    \vspace{-2mm}
    \item When two colors take almost the same coverage and appear to be at a similar distance, we make a judgment call on the color that has more coverage (Fig.~\ref{fig:label_phil} (c)).
    \vspace{-2mm}
    \item When multiple colors appear in the background and none is significantly larger than the rest (Fig.~\ref{fig:label_phil} (d)), or when the object takes up almost all the space in the picture so that we cannot judge the color in the background, or when the perceived color does not belong to our categories, we put it in the ``others'' category.
\end{enumerate}
\begin{algorithm}[t]
        \caption{FlowAug-Gaussian Global $z$}\label{alg:gaussian}
        \label{alg}
            \KwInput{Flow: $\mathcal{F}$, Dataset: $X$, $L$, $\mu, \sigma, b$}
            \For{$l = 1, ..., L$}
            {  
               $x\sim X$\tcp*{Sample image}
               $z,\nu \leftarrow \mathcal{F}_{enc}(x)$ \tcp*{Encode the image}
               $\epsilon \sim \mathcal{N}_{trunc}(\mu,\sigma^2;b) $\tcp*{Sample perturbations}
               $z \leftarrow z + \epsilon $\tcp*{explore space} 
               $x_{aug} \leftarrow \mathcal{F}_{dec}(z,\nu)$ \tcp*{Decode global and local back to image space}
            }
            \KwOutput{$X_{aug}$}

\end{algorithm} \vspace{-4mm}

\subsection{Algorithms}
Knowing properties of $\nu$ and $z$ discussed in \S\ref{sec:flow_property}, we design two families of transformations to operate on global $z$: (1) $T_1$: we add perturbations to $z$, and (2) $T_2$: we interpolate global information extracted from different images.
The over-arching rationale behind is: by equipping models with label-preserving images under diverse environments (i.e., backgrounds), the model should learn more robust correlations\cite{Arjovsky2019InvariantRM}. The second and third row of Fig.~\ref{fig:augmentations} demonstrate our method ability in this regard.



More specifically, in $T_1$ we add truncated Gaussian perturbation $\epsilon$ to $z$,
\vspace{-3mm}
\begin{multline*}
    T_1 :=\{t(x)=\mathcal{F}_{dec}(z+\epsilon,\nu)|(z,\nu)=\mathcal{F}_{enc}(x), \\ \epsilon\sim\mathcal{N}_{trunc}(\mu,\sigma^2; b),~~\forall x\}.
\end{multline*} instead of a Gaussian noise, since a Gaussian noise may sample large numbers that potentially destroy the decoding of $\mathcal{F}_{dec}(z,\nu)$. 
For $T_2$, we decode two random images $x_1, x_2$ to retrieve $z_1, z_2$ and then interpolate $z_1$ and $z_2$ stochastically with a parameter $m$ drawn from a Beta distribution,   
\vspace{-3mm}
\begin{multline*}
    T_2:=\{t(x_i)=\mathcal{F}_{dec}(z_{new},\nu)| z_{new} = m z_i + (1-m) z_j, \\
    m\sim Beta(\alpha,\alpha), (z_i,\nu_i)=\mathcal{F}_{enc}(x_i), ~~\forall i\neq j\}.
\end{multline*}
Detailed transformations are elaborated in Algorithm~\ref{alg:gaussian} \&~\ref{alg:mix}.

\begin{algorithm}[t]
    \caption{FlowAug-Mix Gloabl $z$}\label{alg:mix}
    \label{alg}
        \KwInput{Flow: $\mathcal{F}$, Dataset: $X$, Threshold: $tr$, $L$, $\alpha$}
        \For{$l = 1, ..., L$}
        {  $x_1, x_2 \sim X$\tcp*{Sample images}
           $z_1,\nu_1 \leftarrow \mathcal{F}_{enc}(x_1)$ \tcp*{Encode $x_1$}
           $z_2,\nu_2 \leftarrow \mathcal{F}_{enc}(x_2)$ \tcp*{Encode $x_2$}
           $m\sim Beta(\alpha,\alpha)$\tcp*{Sample interpolation parameter}
           \If{$m < tr$}{
                  $m\leftarrow1-m$ \tcp*{Avoid drastic change in $style$}
           }
           $z_1 \leftarrow m z_1 + (1-m) z_2$\tcp*{explore space} 
           $x_{aug} \leftarrow \mathcal{F}_{dec}(z_1,\nu_1)$ \tcp*{Decode global and local}
        }
        \KwOutput{$X_{aug}$}
        
\end{algorithm}\vspace{-1mm}

We train our models with the following learning objectives: (1) training only with transformed images from $T_1$ or $T_2$ instead of the original examples, (2) in addition to transformed images, adding the original dataset, and (3) combining the two algorithms and the original dataset,
\vspace{-1mm}
\begin{equation}\label{eq:single}
    \mathcal{L}_{FlowAug} = \mathcal{L}(f(t(x)),y; \theta),~~ t\sim T_1~~ or ~~t \sim T_2, 
\end{equation}
\vspace{-3mm}
\begin{multline}\label{eq:single_std}
    \mathcal{L}_{FlowAug+std} = \mathcal{L}(f(t(x)),y; \theta) + \lambda\mathcal{L}(f(x),y; \theta), \\~~ t\sim T_1~~ or ~~t \sim T_2, 
\end{multline}
\vspace{-3mm}
\begin{multline}\label{eq:gauss_std_mix}
    \mathcal{L}_{combine} = \mathcal{L}(f(t_1(x)),y; \theta) + \lambda_1 \mathcal{L}(f(t_2(x)),y; \theta) + \\
    \lambda_2 \mathcal{L}(f(x),y; \theta), t_1\sim T_1~~and~~t_2 \sim T_2,
\end{multline}


\subsection{Quantifying subgroup discrepancy}\label{sec:metric}

\begin{figure*}[t]
     \centering
     \includegraphics[width=\textwidth]{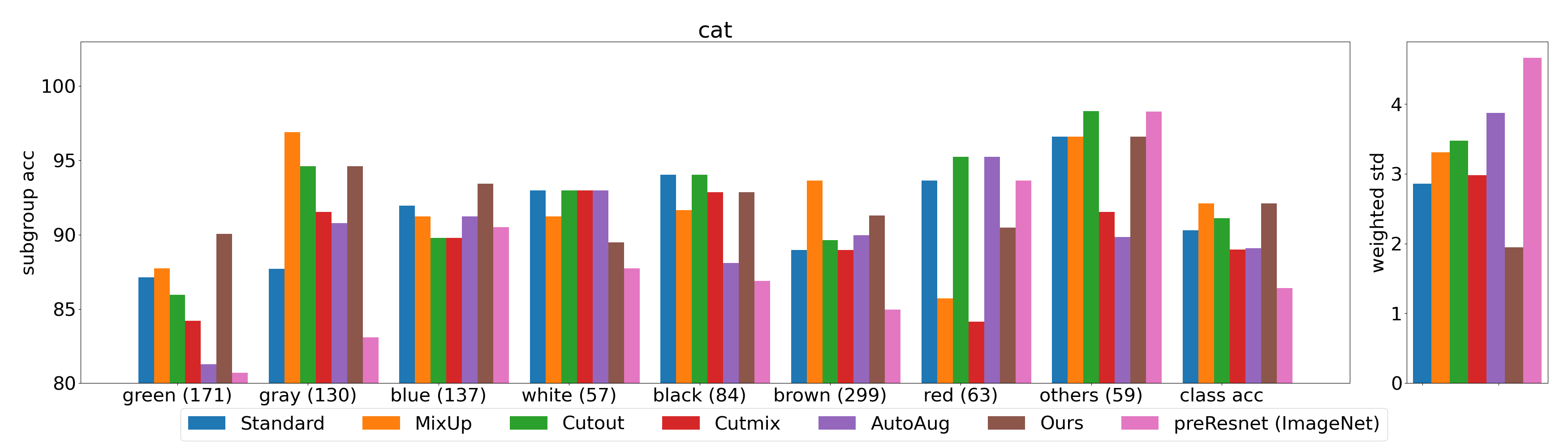}
     \vspace{-7mm}
        \caption{\textbf{Subgroup performances (CIFAR10-Cat).} FlowAug has more balanced results across subgroups and lower WeightedStd, suggesting our method is more resistant to spurious correlations such as background color. (.) indicates the number of instances in the subgroup.\ming{added preResnet18}}
        \label{fig:cat_acc}
        \vspace{-5mm}
\end{figure*}

To quantify the reliance on background attributes, we first propose using the weighted standard deviation, 
\vspace{-1mm}
\begin{equation}\label{eq:wstd}
    \sigma_w = \sqrt{\frac{\sum_{i=1}^G w_i (s_i-\Bar{s}^*)^2}{\sum_{i=1}^G w_i -1 } },
\end{equation}
where $s_i$'s are the subgroup accuracies, $\Bar{s}^*$ the weighted mean, $w_i$'s the weights determined by the number of examples in the subgroup, $G$ the number of groups. Weighted Std can be applied to subgroups performances within a class (as in Fig.~\ref{fig:cat_acc}), and across all accuracies from different classes and subgroups. 

The second metric we propose is macro standard deviation (\emph{MacroStd}),
\vspace{-1mm}
\begin{equation}\label{eq:macrostd}
    \sigma_{Macro}= \sqrt{\frac{1}{C}\sum_{i=1}^{C}{\sigma_{w}^{(i)}}^2}, 
\end{equation}
where $\sigma_w^{(i)}$ is the weighted standard deviation for each class, and $C$ is the number of classes.

\emph{MacroStd} treats each class equally and measures the sensitivity of a model performance across classes. If \emph{MacroStd} is high, this suggests the model has imbalanced performances across classes and also could be affected by background colors. We conduct a correlation analysis to show our metric is a better indicator for both sensitivity and accuracy (see Appendix)\ming{added}.


\vspace{-1mm}
\section{Experiments}\label{sec:experiment}

In this section, we discuss our empirical results on the study of spurious correlation with our CIFAR10-B \& CIFAR100-B and their integration with OOD datasets such as CIFAR10-C and CIFAR100-C. Secondly, although not our primary foci, we present ID and OOD image classification experiments on three datasets --- CIFAR10, CIFAR100, CIFAR10.1 --- to test the generalization capabilities of applying FlowAug. Lastly, we analyze and provide intuitions on how our approach is superior. Furthermore, the comparing baselines and implementation details are provided. 

Due to human resource limit, we are not able to scale labeling efforts to larger benchmark such as ImageNet, but our work has pinpointed critical issues in the subgroup discrepancy in image classification. And we reasonably believe the phenomenon will persist in other datasets given the result from preResnet (Fig.~\ref{fig:case_study} (\textit{right})). In addition, a recent benchmark work \cite{yang2022openood} shows that CIFARs are \emph{not necessarily easier than ImageNet}, which also validates our efforts. \ming{moved here}

\vspace{-1mm}
\subsection{Datasets}  
Other than our CIFAR10-B and CIFAR100-B that are based on CIFAR10 and CIFAR100 \cite{Krizhevsky2009LearningML}, we integrate them with CIFAR10-C \& CIFAR100-C \cite{hendrycks2019robustness}, which are benchmark datasets to model generalization abilities in the presence of 18 shallow corruptions including blurring, contrast, shift, etc. 
Finally, CIFAR10.1 \cite{recht2018cifar10.1} is a test set consists of 2000 images collected from TinyImages \cite{torralba2008tinyimages} and contains the same class labels as CIFAR10. Additionally, we include ImageNet-10 based on our labeling method and discuss the results in the Appendix.

\vspace{-1mm}
\subsection{Baselines}\label{sec:baselines}
We compare our proposed method with four types of low-level DA methods (1) mixing by interpolations, (2) fill-in-with-blank, (3) mixing by fill-in-the-blank, (4) combinations of image manipulations. In our experiments, we compare with the best setups reported in their papers. We include more discussion on rationale behind the selecting the chosen baselines and additional comparisons with composite data augmentations such as AugMix and AugMax in the Appendix.
\vspace{-3mm}
\paragraph{Mixup}~\cite{Zhang2018mixupBE} does linear interpolation on two random images $x_1, x_2$ and mix them as $x_{new}=\lambda x_1 + (1-\lambda) x_2, $ where $\lambda\sim Beta(\alpha,\alpha)$, and the same applies to the label, $y_{new}=\lambda y_1 + (1-\lambda) y_2$.
\vspace{-5mm}
\paragraph{Cutout}~\cite{Devries2017ImprovedRO} randomly crops out a portion of an image and fills it with a specific color, and the label remains unchanged.
\vspace{-5mm}
\paragraph{Cutmix}~\cite{Yun2019CutMixRS} crops out an area of image, but fills the area with a portion of the same size from another image. The label of the augmented image is adjusted according to the proportion of the area of two engaging examples.
\vspace{-5mm}
\paragraph{Autoaug}~\cite{Cubuk2018AutoAugmentLA} uses reinforcement learning to optimize a pre-defined set of policies, combinations of low-level image manipulation, and then learns the best policy for DA.
\vspace{-5mm}
\paragraph{Standard} refers to the models trained on the original datasets, without using any DA methods.

\vspace{-1mm}
\subsection{Implementation Details}

\paragraph{Generative models} We pre-train the normalizing flow models as in \cite{decoupling2021}, and they achieve the negative log-likelihood scores in bits/dim (BPD) 3.27 and 3.31 on CIFAR10 and CIFAR100, respectively. 
\vspace{-3mm}
\paragraph{Hyperparameters} In Algorithm~\ref{alg:gaussian}, we simply set $\mu=0$ and $\sigma=0.1$ for the truncated Gaussian distribution. As for truncation $b$, we empirically find that $z$ has an average maximum value around 4 and so we set $b=4$. In Algorithm~\ref{alg:mix}, we simply set $\alpha=1$ and $tr=0.5$. For all models reported in Table~\ref{tab:acc}, we train Resnet18 for 250 epochs with weight decay 0.0005. Also, the learning rate starts at 0.1 and is divided by 10 at [100, 150] epochs. For our learning objectives, we lightly fine-tune $\lambda$ in Eq.~\eqref{eq:single_std} with values of $\{0.01, 0.05, 0.1\}$, and $\lambda_1, \lambda_2$ in Eq.~\eqref{eq:gauss_std_mix} with $\lambda_1=1$ and $\lambda_2 \in\{0.01, 0.05, 0.1\}$.
The generative flow models are trained on two NVIDIA A40 GPUs, while the Resnet18 are trained on one NVIDIA A40 GPU.

\subsection{Empirical Results}
\paragraph{MacroStd and WeightedStd} Table~\ref{tab:var} reports the \emph{MacroStd} and the weighted standard deviation of subgroup performances from the whole dataset. Our approach consistently has both lower \emph{MacroStd} and lower WeightedStd over the baselines. Moreover, in Fig.~\ref{fig:cat_acc}, our approach also achieves lower WeightedStd at the class level. These results show  evidence that our approach is less affected by the background colors and hence is more robust. 

\begin{table}
 
 \centering
  \begin{adjustbox}{max width=1\linewidth}
  \begin{tabular}{lcccc}
    \toprule
       &  \multicolumn{2}{c}{MacroStd} &  \multicolumn{2}{c}{Weighted Std}\\
       & CIFAR10 & CIFAR100 & CIFAR10 & CIFAR100 \\
     \midrule
    Standard & 2.24 & 12.24 & 3.45 &16.44\\
    Mixup    & 2.17 & 11.94 &3.02 &16.75\\
    Cutout &  1.91 & 12.45  &2.94 &16.52\\
    Cutmix &  1.91 & 12.62  &3.34 &16.87\\
    AutoAug & 2.11 & 11.74  &3.54 &16.30\\
    \midrule
    
    Ours (Mix $z$) & 1.99 & 11.73 & 2.92 &15.96\\
    + Std & 1.81 & 12.49  & 2.76 & 16.76\\
    Ours (Uniform on $z$)  & 1.83 & 12.17 & 2.98 & 16.57\\
    + Std           &  1.81  &   11.59 & 3.26 & 15.95 \\
    + Mix $z$       & 1.85 & 11.72 & 2.89 & 16.31  \\
    + Std + Mix $z$ & 1.86 &  \textbf{11.23} & 3.12 & 16.01 \\
    Ours (Trunc Gaussian on $z$) & 1.91 & 12.00 & 2.82 &16.23\\
    + Std & \textbf{1.65} & 11.55 & 3.08 & 16.02 \\
    + Mix $z$       & 1.89 & 11.95 & 3.09 & 16.35 \\
    + Std + Mix $z$ & 1.66 & 11.78 & \textbf{2.71} &\textbf{15.92}\\
    \midrule
    Trunc Gaussian on $\nu$ & 1.94 & 12.68 & 2.81 &16.70\\
    Mix $\nu$ & 2.50 & 13.27 & 4.21& 18.46\\
    \bottomrule
  \end{tabular}
  \end{adjustbox}
  \vspace{-2mm}
  \caption{\textbf{\emph{MacroStd} and Weighted Std.} Lower numbers represent lower reliance on spurious background color correlations and our algorithms are consistently better than the baselines.\ming{added new exp results in L10 \& L14}}
  \vspace{-2mm}
  \label{tab:var}
\end{table}

\begin{figure}[t]  
     \centering
         \includegraphics[width=1\linewidth]{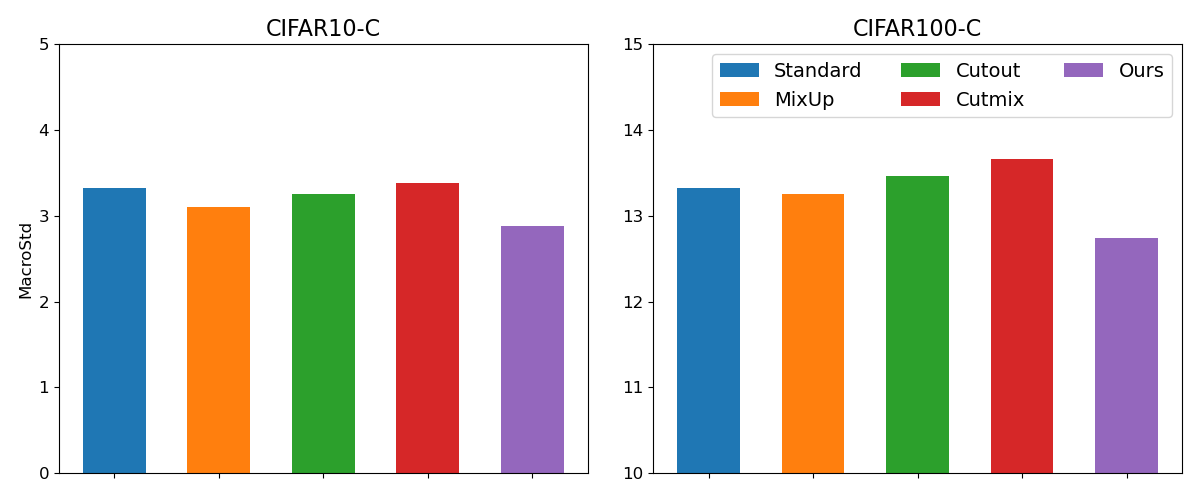}
         \vspace{-5mm}
     \caption{\textbf{CIFAR10-C and CIFAR100-C results (severity=1).} We integrate our datasets with CIFAR10/100-C. FlowAug demonstrates more consistent subgroup performances on OOD datasets.}
     \vspace{-5mm}
     \label{fig:cifarc_std}
\end{figure}

\vspace{-2mm}
\paragraph{CIFAR10-C and CIFAR100-C} \ming{new section} Another benefit of our datasets is the compatibility with CIFAR-Cs. Together with CIFAR-Cs we are able to evaluate the \emph{subgroup discrepancy} phenomenon in an OOD setting. Fig.~\ref{fig:cifarc_std} shows that FlowAug has reduced \emph{subgroup discrepancy} than other DAs. We exclude AutoAug in Fig.~\ref{fig:cifarc_std} because it contains policies resembling some corruption types of CIFAR-C so we deem it not a fair comparison.

\vspace{-3mm}
\paragraph{CIFAR10 and CIFAR100} Although ID and OOD generalization performances are not our main foci, our FlowAug demonstrates significant gains on CIFAR10 and CIFAR100 and we report our experimental results in Table~\ref{tab:acc}. Algorithm~\ref{alg:gaussian} itself achieves results better than all the baselines.  Algorithm~\ref{alg:mix} also performs better than the \emph{Standard} baseline and is competitive with other methods. 
When Algorithm~\ref{alg:gaussian} and \ref{alg:mix} are combined or also add the \emph{Standard} loss (Eq.~\eqref{eq:gauss_std_mix}), they can further enhance the performances. 

In Algorithm~\ref{alg:gaussian}, we do an ablation study with uniform distribution in Sec.~\ref{sec:gaussain_uniform}. The best improvements on CIFAR10 and CIFAR100 are at 1.42\% and 1.47\% respectively. The superior results of our deep generative augmentation approach with decoupled representations have shown greater generalization potential. 

\begin{table}
  
  \begin{adjustbox}{max width=\linewidth}
  \begin{tabular}{lccc}
    \toprule
     & CIFAR-10 & CIFAR10.1 & CIFAR-100 \\
     & best / last & best / last & best / last \\
     \midrule
 Standard & 95.16 / 95.00   & 88.70 / 88.25 &  78.52 / 78.52 \\
 Mixup & 95.96 / 95.82   & 89.75 / 88.85 & 77.91 / 76.99 \\
 Cutout & 95.94 / 95.60   & 90.40 / 89.70 & 78.21 / 78.00 \\
 Cutmix & 96.17 / 96.04  & 90.05 / 90.20 & 78.87 / 78.42 \\
 AutoAug & 95.82 / 95.47 & 89.85 / 89.80 & 78.57 / 78.12\\
 \midrule
 Ours (Mix $z$)                  & 95.98 / 95.59 & 90.35 / 89.30 & 78.58 / 78.07 \\
  + Std                          & 96.15 / 95.73 & 90.25 / 90.20 &  78.96 / 78.33\\
 Ours (Uniform on $z$)           & 96.12 / 96.11 & 90.35 / 90.45 &  79.45 / 79.24\\
 + Std           &  96.28 / 96.14 &  90.95 / 91.15 & \underline{79.68} / \textbf{79.67} \\
 + Mix $z$                       & \textbf{96.58} / \textbf{96.37}  & \textbf{91.40} / \textbf{91.65} & \underline{79.68} / 79.52  \\
 + Std + Mix $z$           & 96.44 / \underline{96.31} & \underline{91.05} / 91.00 &  \underline{79.68} / 79.51 \\
 Ours (Trunc Gaussian on $z$)    & 96.23 / 96.23 &  90.15 / 90.25 & 79.30 / 79.04 \\
  + Std                          & 96.22 / 96.04 &  90.55 / 89.80 &  79.62 / \underline{79.62}\\
  + Mix $z$                      & 96.49 / 96.42  & 90.05 / 90.25 &  79.51 / 79.18 \\
 + Std + Mix $z$                 & \underline{96.53} / 96.29&  90.75 / \underline{91.20} & \textbf{79.99} / 79.54\\
 \midrule
 Trunc Gaussian on $\nu$ & 95.70 / 95.45 &  88.70 / 89.25 &  78.50 / 78.41\\
 Mix $\nu$ & 94.22 / 93.43 & 87.55 / 85.50 & 74.05 / 71.42 \\
    \bottomrule
  \end{tabular}
  \end{adjustbox}
  \vspace{-2mm}
  \caption{\textbf{Test results in \% (best/last epoch).} Although ID and OOD generalization are not our foci, FlowAug consistently outperforms the baseline and we only highlight the top-2 results. Note: we simply apply the models trained from CIFAR-10 to obtain CIFAR10.1 results without fine-tuning. \ming{added new exp results in L10 \& L14}}
  \vspace{-5mm}
  \label{tab:acc}
\end{table}

\vspace{-3mm}
\paragraph{CIFAR10.1} On another OOD dataset CIFAR10.1, we also observe significant improvements in performances from FlowAug over the baselines (up to +1\% better than the best of all five baselines). These results again demonstrate that FlowAug is more robust and has better generalizability.

Our CIFAR10.1, CIFAR10-C, and CIFAR100-C experiments demonstrate FlowAug's generalizability to OOD data and validate our approach of using deep decoupled representations for DA.

\vspace{-1mm}
\subsection{Analysis}
Our two FlowAug algorithms~\ref{alg:gaussian} \& \ref{alg:mix} both improve over the baselines, and combining the two shows even superior results. Conceptually, we know that the flow model $\mathcal{F}$ can map $X$ to a Gaussian prior distribution (cf. Eq.~\ref{eq:encoder_decoder}), but not necessarily all the points in the Gaussian distribution would follow the reverse $g$ to a realistic image. Then intuitively, given that $z_1, z_2$ come from real images, Algorithm~\ref{alg:mix}'s interpolating of $z_1, z_2$ can be interpreted as finding an optimal point between two proven optimal points in the space, i.e., Algorithm~\ref{alg:mix} explores the Gaussian space in an efficient way.

On the other hand, adding a sampled perturbation to $z$ as in Algorithm~\ref{alg:gaussian} can stretch the search space to outside of the Gaussian, which brings good performances. It also explains why a combined approach such as Eq.~\ref{eq:gauss_std_mix} can generally achieve superior performances over the rest since Algorithm~\ref{alg:gaussian} and \ref{alg:mix} can be complementary.

\vspace{-1mm}
\section{Ablation Studies}
To further study global and local representations, we can make some design choices applied to $z$ and $\nu$. In \S~\ref{sec:flowaug}, we assume $z$ and $\nu$ carry information about the background and ground truth respectively, and we want to test the assumptions and the generality of Algorithm \ref{alg:gaussian}.

\vspace{-1mm}
\subsection{Perturbing Local ($\nu$) or Global ($z$)?} \label{sec:ablation_loc_v_global}
\S~\ref{sec:experiment} has shown that perturbing $z$ improves generalization and the robustness of models. On the other hand, we can also decode realistic images by perturbing $\nu$ (Fig.~\ref{fig:augmentations}(h)), which we assume affects the prediction. We apply Algorithm~\ref{alg:gaussian} \& \ref{alg:mix} with the same parameters on $\nu$, and the results deteriorate on all datasets by at least 0.5\% and up to 7\%, suggesting that our assumption about $\nu$'s correspondence to the ground truth label is reasonable.

\vspace{-1mm}
\subsection{Does perturbation type matter? A case study of Gaussian vs Uniform distributions}\label{sec:gaussain_uniform} Algorithm~\ref{alg:gaussian} uses a truncated Gaussian perturbation, but in fact, we can also add noise sampled from other distributions, such as a Uniform perturbation. To have about the same amount of probability density in the same range as $\mathcal{N}(\mu=0,\sigma=0.1)$, we choose $\mathcal{U}(-0.2,0.2)$ for our study. Table~\ref{tab:acc} shows that adding uniform noise is comparable to adding truncated Gaussian, and when combined with algorithm~\ref{alg:mix} and(or) \emph{Standard}, the improvements are top-2, achieving over a 1\% gain on CIFAR10 and CIFAR100, and more than a 2\% gain on CIFAR10.1. This study suggests the generalization capability of FlowAug on symmetric noise distributions. 



\subsection{Can FlowAug be composited with another method?}

Composite DA sometimes offer additional benefit to generalization \cite{wang2021augmax,Hendrycks2020AugMixAS,vryniotis2021train}. Thus, we investigate if FlowAug possesses the flexibility of being composited and run an additional experiment combining our simplest Algorithm \ref{alg:gaussian} variation and Cutmix. Tab.\ref{tab:appx_alg1_cutmix} shows combining FlowAug and Cutmix further mitigates subgroup degradation and also enhances generalization. In addition, on CIFAR100, the test accuracy is better than any methods using Resnet18 to our knowledge. This experiment showcases the possibility of chaining FlowAug with other methods for attaining ``SOTA'' performances in both mitigating bias or enhancing generalization.  We compare FlowAug with other recent composite DA such as AugMix, AugMax in the Appendix. 
\begin{table}
  \centering
  \begin{adjustbox}{max width=\linewidth}
  \begin{tabular}{lccc|ccc}
    \toprule
     & CIFAR10 & MacroStd & W-Std & CIFAR100 & MacroStd & W-Std\\
     \midrule
    Standard  & 95.16 & 2.24 & 3.45 & 78.52 & 12.24 & 16.44\\
    Cutmix & 96.17 & 1.91 & 3.34 & 78.87 & 12.62 & 16.87\\
    Ours-alg1 & \underline{96.23} & \underline{1.83} & \underline{2.98} & \underline{79.45} & \underline{12.17} & \underline{16.57} \\
    Ours-alg1+Cutmix & \textbf{96.34} & \textbf{1.65} & \textbf{2.91} & \textbf{81.57} & \textbf{11.20} & \textbf{15.34}\\
    \bottomrule
  \end{tabular}
  \end{adjustbox}
  \caption{\textbf{Chaining FlowAug with Cutmix on CIFAR10/100.} Our FlowAug has the flexibility of being combined with other works to further enhance performances.}
  \vspace{-4mm}
  \label{tab:appx_alg1_cutmix}
\end{table}

\vspace{-1mm}

\section{Correlation Analysis on \textit{MacroStd} and Performances} 

WeightedStd is the most common measure to quantify sensitivity in statistics. However, we want to justify our MacroStd to be a more suitable metric in quantifying subgroup degradation. We perform correlation analyses between accuracy and MacroStd(ours)/WeightedStd. The statistics are summarized in Table~\ref{tab:correlation}, and on both CIFAR10 and CIFAR10.1, our metric is a better indicator for sensitivity and accuracy (coefficients the lower the better), which validates the novelty of our metric.

\begin{table}[h]
\centering
  \begin{adjustbox}{max width=1\linewidth}
  \begin{tabular}{lcc}
    \toprule
     & CIFAR-10 & CIFAR10.1  \\
     & best / last & best / last \\
     \midrule
    MacroStd (ours) & -0.89 / -0.85   & -0.83 / -0.88\\
    WeightedStd & -0.78 / -0.77 &   -0.62 / -0.70 \\
    \bottomrule
  \end{tabular}
  \end{adjustbox}
  \caption{\textbf{Correlation ($\downarrow$) between accuracy and sensitivity metrics (best/last epoch).}}
  \vspace{-5mm}
  \label{tab:correlation}
\end{table}

\section{Related Works}

\paragraph{Representation Learning.} Deep learning models' success is generally attributed to their ability to learn complex and meaningful representations~\cite{6472238}, and most attempts to learning quality representations require certain inductive biases, for instance, space invariance of CNNs \cite{726791}. Of particular interest to our work, generative models such as VAEs \cite{Burgess2018UnderstandingDI,Mathieu2018DisentanglingD, Chen2018IsolatingSO, Ding2020GuidedVA} enforce constraints such as independent multivariate Gaussain in the latent layers to learn disentangled representations. Our work leverages a model that learns two decoupled representations instead of the factorial ones. 

\vspace{-3mm}
\paragraph{Data Augmentation.} 
DA often helps achieve improved generalization. One line of approach performs low-level basic image operations  such as mixing examples \cite{Zhang2018mixupBE}, or random erasing \cite{Yun2019CutMixRS, Devries2017ImprovedRO}, etc. Another approach uses reinforcement learning to learn the best policy of basic image operations \cite{Cubuk2018AutoAugmentLA, Cubuk2020RandaugmentPA}. \cite{Mao2021GenerativeIF} use causal inference to guide their method and add interventions during the generation process. Our work uses decoupled global representations to isolate spurious correlations and then learn robust correlations to the objects. We refer readers to \cite{Feng2021ASO} for recent surveys.

\vspace{-3mm}
\paragraph{Robustness.} Robustness in DNNs has drawn the attention of the community largely since \cite{Goodfellow2015ExplainingAH,Kurakin2017AdversarialML}. Multiple lines of research were proposed to study robustness, including using Distributionally Robust Optimization \cite{duchi2020learning,10.2307/23359484}, Adversarial Training \cite{Madry2018TowardsDL, chen2021depois,DBLP:journals/corr/abs-2003-01908,NEURIPS2018_358f9e7b}, and certifiable bounds \cite{DBLP:journals/corr/abs-1906-06316,pmlr-v97-cohen19c}. \cite{Arjovsky2019InvariantRM} proposed a scenario where representations learned should be robust in different environments, and \cite{Scholkopf2021TowardCR, Scholkopf2022FromST} suggests that learning Causal Representations can be an ultimate approach to robustness in deep learning. \cite{geirhos2018imagenet} studied the effect of shape and texture to CNNs. Our work is in line with the idea of \cite{Arjovsky2019InvariantRM} and studies color as a spurious factor orthogonal to concept of texture \cite{bianconi2021colour}.

\vspace{-1mm}
\section{Conclusion and Future Work}

\ming{re-written} In this work, we contributed 20,000 non-trivial human annotations in two datasets to reveal the phenomenon of subgroup discrepancy in various (pre)training techniques, and proposed a semantic DA method, \textbf{FlowAug} which trains more robust models evaluated on CIFARs and CIFAR-Cs. Additionally, we showed the potential of using disentangled representations for DA by achieving superior generalization performances on both ID and OOD datasets.

\ming{re-written.}We believe our work serves as a leap forward in studies of fairness, robustness, and even causality in DNNs, as we can use CIFAR-Bs to quantify the effect of a hidden bias and we learn that high-level DA is suited to achieve consistent predictions. Last but not least, due to human and computing resource limits, we are not able to scale the labeling effort nor the experiments to larger datasets such as ImageNet, but our results on ImageNet-10 and CIFARs have shown it is an impactful direction and should further enhance the performances. 


{\small
\bibliographystyle{ieee_fullname}
\bibliography{egbib}
}

\newpage
\onecolumn
\appendix
\section{CIFAR10-B Statistics}

\begin{table*}[h]
\centering
  \begin{adjustbox}{max width=\textwidth}
   \begin{tabular}{lcccccccc}
    \toprule
    Classes & Green & Gray & Blue & White & Black & Brown & Red & Others \\
     \midrule
    cat & 171 & 130 & 137 & 57 & 84 & 299 & 63 & 59 \\
    dog & 271 & 90 & 124 & 39 & 81 & 297 & 58 & 40 \\
    truck & 121 & 180 & 278 & 187 & 24 & 177 & 2 & 31 \\
    bird & 453 & 58 & 179 & 32 & 32 & 237 & 0 & 9 \\
    airplane & 85 & 112 & 609 & 81 & 7 & 102 & 4 & 0 \\
    ship & 71 & 88 & 711 & 45 & 15 & 56 & 0 & 14 \\
    frog & 417 & 64 & 75 & 56 & 49 & 238 & 15 & 86 \\
    horse & 493 & 29 & 108 & 41 & 50 & 251 & 7 & 21 \\
    deer & 604 & 27 & 99 & 8 & 25 & 227 & 4 & 6 \\
    automobile & 177 & 262 & 147 & 130 & 34 & 230 & 15 & 5 \\
    \bottomrule
  \end{tabular}
  \end{adjustbox}
  \caption{\textbf{CIFAR10-B Statistics.} The number of instances that belongs to each background color from each class.}
  \label{tab:cifar10b_appendix}
\end{table*}

\vspace{-8mm}
\section{More discussion on chosen baselines and comparison to AugMix \& AugMax}

We choose the DAs that are strong and commonly studied in recent works such as \cite{wightman2021resnet,touvron2021training,vryniotis2021train}. PyTorch's official ``SOTA'' training method \cite{vryniotis2021train} also uses the combinations of our chosen baselines along with other perfected hyperparameters such as longer training; however, we isolated each baseline to measure the individual contribution of a single method. In addition, we compare with two recent works AugMix and AugMax, which are recent methods evaluated on plain Resnet18, and our generalization results are superior, as per reported in \cite{wang2021augmax} (see Tab.\ref{tab:augmix_n_inet10}-\textit{left}). AugMix and AugMax are not a single method but a combinations of different ones and were designed for other robustness purposes. Lastly, we want to re-iterate that generalization performance is not our main focus but an additional benefit when reducing the model bias toward specific colors.

\begin{table}
  \centering
  \begin{adjustbox}{max width=\linewidth}
  \begin{tabular}{lcc|}
    \toprule
     & CIFAR10 & CIFAR100 \\
     \midrule
    AugMix  & 95.79 & 78.23 \\
    AugMax & 95.76 & 78.96 \\
    Ours & \textbf{96.58} & \textbf{79.99} \\
    \bottomrule
  \end{tabular}

  \begin{tabular}{lccc}
    \toprule
     & Acc. & MacroStd & W-Std \\
     \midrule
    Standard  & 91.80 & 9.57 & 11.17\\
    Autoaug & 93.40 & 8.84 & 9.99\\
    Ours-Alg1 & \textbf{94.80} & \textbf{8.39} & \textbf{9.96}\\
    \bottomrule
  \end{tabular}
  
  \end{adjustbox}
  \caption{\textbf{\textit{Left}: Accuracy of our FlowAug compared with AugMix \& AugMax. \textit{Right}: ImageNet-10 results.} Our FlowAug show superior results.}
  \label{tab:augmix_n_inet10}
\end{table}

\section{ImageNet-10}
We labeled ImageNet-10 and found similar subgroup degradation aspect on ImageNet, and our method can still mitigate the issue using standard training procedure without fine-tuning (Tab.\ref{tab:augmix_n_inet10}-\textit{right}); therefore, we believe FlowAug will work on high-resolution/larger datasets

\section{Selection of Color Groups\ming{new section}}
While labeling the datasets, we add one color only if it has a couple of images. This prevents having many colors with 0 images. For example, in CIFAR10, classes like ``bird" and ``ship" do not contain images with a red background, but it has some presence in other classes, so we added red as a color group.

\section{Generalization on CIFAR10-C and CIFAR100-C}
\begin{figure}[t]
     
     \centering
         \includegraphics[width=0.5\linewidth]{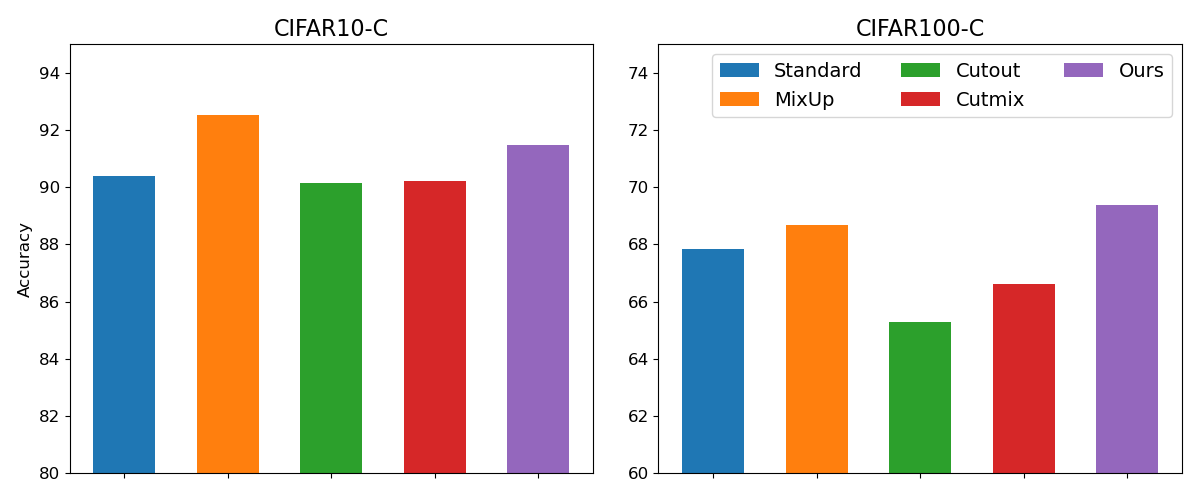}
         
     \caption{\textbf{CIFAR10-C and CIFAR100-C results (severity=1).} FlowAug's results are comparable with common DA methods that have corruption effects such as \emph{Mixup}, even though FlowAug does not add corruptions to training. Note that \emph{Mixup} (Figure~\ref{fig:augmentations}(b)) produces a similar effect to blurrings.
     }
     \vspace{-5mm}
     \label{fig:cifarc}
\end{figure}

Aside from generalization on \emph{i.i.d.} data, we are interested in FlowAug's capabilities to generalize to out-of-domain (OOD) data, which is another aspect of robustness. We use models of the last epoch from Table~\ref{tab:acc} to test on CIFAR10-C and CIFAR100-C. Although FlowAug does not explicitly add corruptions such as various kinds of blurring, contrast and so on to training data, we observe comparable performances (Fig.~\ref{fig:cifarc}) with augmentation methods that have corruption effects, such as \emph{MixUp}, and FlowAug is better than \emph{Cutout} and \emph{Cutmix} (+1.35\% and +1.26\%, respectively). 

     
         

\section{Switch Operation}

\begin{figure}[h!]
     \centering
     \includegraphics[width=0.5\textwidth]{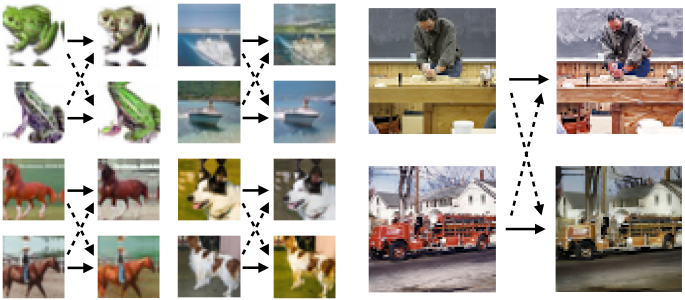}
     \vspace{-3mm}
        \caption{\textbf{Examples of switch operation on decoupled representations.} \cite{decoupling2021} can perform \emph{switch} operations on global and local representations of images on various datasets (figure used with the author's permission).}
        \vspace{-8mm}
        \label{fig:switch}
\end{figure}

\section{Label Quality\ming{new section}}
The background color labels are labeled by a person with an experienced computer vision background for consistency and are verified twice. As a quality check, two people with strong technical backgrounds checked 500 random images. The rate of agreement is 91.8 percent, and 5.4 percent of images that did not agree in the first round agree with the labels we used for experiments after discussion. The disagreement rate is smaller than the average error rate in modern datasets \cite{northcutt2021pervasive}. We will release the dataset and welcome the community to update the background labels.

\vspace{-2mm}
\section{An Additional Protected Attribute\ming{new section}}
\begin{figure}[h]  
     \centering
         \includegraphics[width=0.5\linewidth]{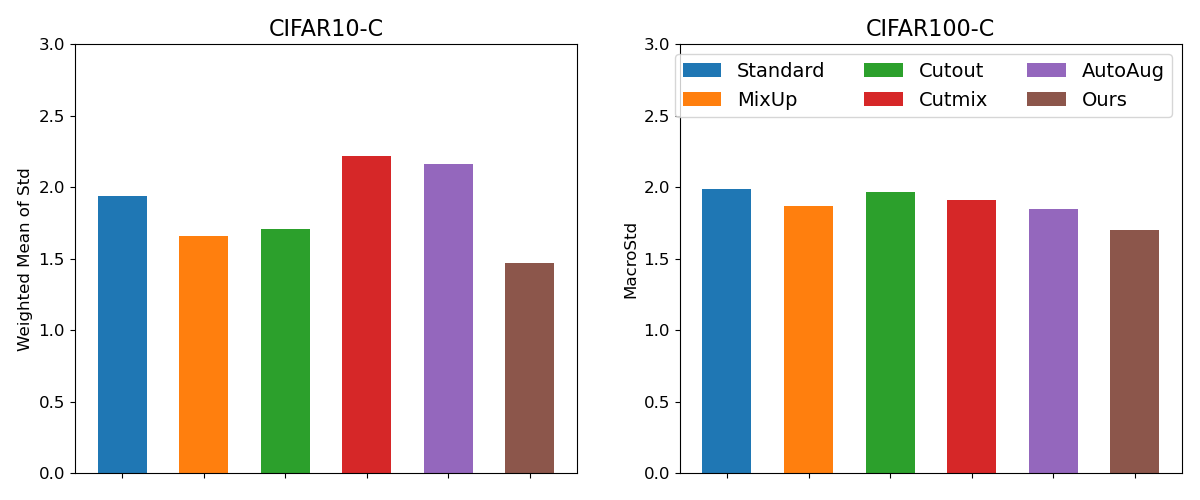}\vspace{-4mm}
     \caption{\textbf{Additional protected attribute on CIFAR10 and CIFAR100.} }
     \vspace{-4mm}
     \label{fig:cifar_additional_protected}
\end{figure}
In this work, we studied color as a bias, and we can also study another protected attribute in a hand-wavy fashion. For example on CIFAR10, we can group the ten classes into vehicles and animals and apply our MacroStd to measure sensitivity. A similar study can be conducted on CIFAR100 with its original 20 super-classes. 
On CIFAR10, since there are only two ``super-classes'' so we report the weighted average of the standard deviation of vehicle/animal group; on CIFAR100, we report our MacroStd across the 20 superclasses. The results are summarized in Fig.~\ref{fig:cifar_additional_protected}, and it shows FlowAug is again superior. We emphasize that this is a conceptual study on an additional protected attribute and is not within the scope of our work.

\section{CIFAR100-B Statistics}

\begin{table*}[h]
\centering
  \begin{adjustbox}{max width=\textwidth}
   \begin{tabular}{lcccccccc}
    \toprule
    Classes & Green & Gray & Blue & White & Black & Brown & Red & Others  \\
     \midrule
    apple & 23 & 6 & 9 & 40 & 11 & 11 & 0 & 0 \\
    aquarium fish & 30 & 5 & 12 & 0 & 38 & 9 & 4 & 2 \\
    baby & 18 & 16 & 25 & 7 & 7 & 21 & 5 & 1 \\
    bear & 56 & 15 & 9 & 1 & 2 & 16 & 1 & 0 \\
    beaver & 36 & 8 & 18 & 8 & 6 & 24 & 0 & 0 \\
    bed & 8 & 21 & 7 & 18 & 3 & 42 & 0 & 1 \\
    bee & 24 & 8 & 10 & 6 & 4 & 33 & 8 & 7 \\
    beetle & 29 & 13 & 8 & 15 & 1 & 28 & 3 & 3 \\
    bicycle & 24 & 30 & 17 & 7 & 3 & 17 & 2 & 0 \\
    bottle & 10 & 24 & 9 & 17 & 7 & 28 & 4 & 1 \\
    bowl & 6 & 19 & 19 & 14 & 22 & 18 & 2 & 0 \\
    boy & 21 & 13 & 16 & 13 & 8 & 21 & 5 & 3 \\
    bridge & 11 & 9 & 67 & 5 & 5 & 2 & 1 & 0 \\
    bus & 14 & 19 & 23 & 18 & 8 & 14 & 0 & 4 \\
    butterfly & 53 & 12 & 3 & 7 & 4 & 17 & 2 & 2 \\
    camel & 31 & 10 & 25 & 3 & 9 & 19 & 2 & 1 \\
    can & 5 & 24 & 12 & 25 & 7 & 27 & 0 & 0 \\
    castle & 4 & 12 & 65 & 15 & 1 & 2 & 0 & 1 \\
    caterpillar & 63 & 6 & 7 & 0 & 6 & 15 & 2 & 1 \\
    cattle & 47 & 5 & 13 & 8 & 5 & 20 & 1 & 1 \\
    chair & 3 & 10 & 4 & 71 & 4 & 6 & 0 & 2 \\
    chimpanzee & 65 & 3 & 4 & 3 & 1 & 20 & 1 & 3 \\
    clock & 3 & 21 & 12 & 34 & 4 & 20 & 1 & 5 \\
    cloud & 0 & 2 & 80 & 2 & 8 & 2 & 2 & 4 \\
    cockroach & 2 & 19 & 13 & 44 & 1 & 14 & 5 & 2 \\
    couch & 5 & 14 & 19 & 24 & 6 & 27 & 3 & 2 \\
    crab & 9 & 23 & 16 & 16 & 15 & 16 & 3 & 2 \\
    crocodile & 39 & 13 & 11 & 2 & 5 & 28 & 2 & 0 \\
    cup & 9 & 27 & 18 & 17 & 14 & 11 & 1 & 3 \\
    dinosaur & 23 & 14 & 12 & 26 & 7 & 17 & 0 & 1 \\
    dolphin & 14 & 11 & 74 & 0 & 0 & 1 & 0 & 0 \\
    elephant & 48 & 6 & 17 & 3 & 3 & 19 & 2 & 2 \\
    flatfish & 15 & 14 & 34 & 13 & 6 & 12 & 6 & 0 \\
    forest & 4 & 7 & 22 & 7 & 0 & 11 & 3 & 46 \\
    fox & 32 & 11 & 19 & 1 & 7 & 26 & 2 & 2 \\
    girl & 15 & 11 & 15 & 10 & 13 & 26 & 7 & 3 \\
    hamster & 11 & 23 & 22 & 4 & 8 & 22 & 9 & 1 \\
    house & 20 & 4 & 42 & 20 & 3 & 7 & 0 & 4 \\
    kangaroo & 44 & 14 & 4 & 2 & 3 & 31 & 1 & 1 \\
    keyboard & 8 & 15 & 23 & 8 & 5 & 23 & 3 & 15 \\
    lamp & 5 & 20 & 23 & 14 & 12 & 20 & 5 & 1 \\
    lawn_mower & 32 & 6 & 6 & 45 & 2 & 9 & 0 & 0 \\
    leopard & 31 & 14 & 20 & 1 & 8 & 22 & 1 & 3 \\
    lion & 36 & 4 & 19 & 1 & 3 & 31 & 3 & 3 \\
    lizard & 13 & 14 & 21 & 5 & 8 & 30 & 5 & 4 \\
    lobster & 14 & 12 & 18 & 17 & 11 & 15 & 6 & 7 \\
    man & 15 & 19 & 14 & 9 & 18 & 24 & 1 & 0 \\
    maple_tree & 11 & 6 & 48 & 29 & 2 & 4 & 0 & 0 \\

    \bottomrule
  \end{tabular}
  \end{adjustbox}
  \caption{\textbf{CIFAR100-B Statistics (part 1).} The number of instances that belongs to each background color from each class.}
  \label{tab:cifar100b_appendix}
\end{table*}

\begin{table*}[h]
\centering
  \begin{adjustbox}{max width=\textwidth}
   \begin{tabular}{lcccccccc}
    \toprule
    Classes & Green & Gray & Blue & White & Black & Brown & Red & Others  \\
     \midrule
     motorcycle & 11 & 28 & 10 & 30 & 2 & 16 & 3 & 0 \\
    mountain & 0 & 7 & 83 & 6 & 1 & 1 & 2 & 0 \\
     mouse & 15 & 13 & 13 & 12 & 6 & 34 & 1 & 6 \\
    mushroom & 57 & 9 & 5 & 3 & 7 & 19 & 0 & 0 \\
     oak_tree & 6 & 2 & 72 & 15 & 1 & 4 & 0 & 0 \\
    orange & 13 & 11 & 17 & 20 & 20 & 6 & 3 & 10 \\
    orchid & 35 & 4 & 7 & 4 & 35 & 9 & 1 & 5 \\
    otter & 28 & 14 & 27 & 3 & 5 & 22 & 1 & 0 \\
     palm_tree & 3 & 7 & 62 & 13 & 5 & 5 & 0 & 5 \\
    pear & 19 & 13 & 10 & 28 & 7 & 20 & 0 & 3 \\
    pickup_truck & 31 & 27 & 16 & 6 & 5 & 14 & 0 & 1 \\
    pine_tree & 7 & 11 & 61 & 13 & 1 & 7 & 0 & 0 \\
    plain & 0 & 13 & 70 & 14 & 0 & 3 & 0 & 0 \\
     plate & 3 & 22 & 15 & 26 & 17 & 13 & 2 & 2 \\
    poppy & 57 & 4 & 7 & 4 & 14 & 4 & 2 & 8 \\
     porcupine & 51 & 11 & 8 & 1 & 11 & 17 & 0 & 1 \\
    possum & 27 & 19 & 10 & 1 & 16 & 22 & 2 & 3 \\
    rabbit & 34 & 6 & 12 & 2 & 16 & 28 & 2 & 0 \\
    raccoon & 30 & 6 & 14 & 5 & 21 & 23 & 1 & 0 \\
    ray & 23 & 11 & 49 & 3 & 6 & 8 & 0 & 0 \\
    road & 47 & 7 & 34 & 2 & 2 & 7 & 1 & 0 \\
    rocket & 7 & 14 & 66 & 5 & 4 & 4 & 0 & 0 \\
    rose & 54 & 3 & 10 & 9 & 11 & 5 & 1 & 7 \\
    sea & 2 & 6 & 74 & 2 & 0 & 12 & 1 & 3 \\
    seal & 17 & 14 & 43 & 4 & 6 & 12 & 1 & 3 \\
    shark & 12 & 3 & 68 & 1 & 13 & 2 & 1 & 0 \\
    shrew & 30 & 8 & 17 & 5 & 5 & 31 & 3 & 1 \\
    skunk & 42 & 13 & 5 & 3 & 6 & 30 & 1 & 0 \\
    skyscraper & 1 & 5 & 80 & 9 & 2 & 2 & 0 & 1 \\
    snail & 45 & 7 & 12 & 3 & 5 & 26 & 1 & 1 \\
    snake & 21 & 15 & 13 & 7 & 4 & 30 & 7 & 3 \\
    spider & 37 & 18 & 15 & 2 & 9 & 11 & 7 & 1 \\
    squirrel & 36 & 6 & 14 & 5 & 8 & 30 & 0 & 1 \\
    streetcar & 22 & 10 & 32 & 11 & 4 & 13 & 7 & 1 \\
    sunflower & 44 & 1 & 26 & 9 & 7 & 3 & 2 & 8 \\
    sweet_pepper & 15 & 7 & 9 & 27 & 14 & 14 & 6 & 8 \\
    table & 10 & 17 & 18 & 17 & 7 & 29 & 2 & 0 \\
    tank & 16 & 13 & 30 & 17 & 1 & 23 & 0 & 0 \\
    telephone & 5 & 11 & 5 & 59 & 4 & 13 & 3 & 0 \\
    television & 4 & 24 & 10 & 26 & 10 & 21 & 2 & 3 \\
    tiger & 45 & 3 & 14 & 8 & 9 & 16 & 0 & 5 \\
    tractor & 33 & 8 & 34 & 10 & 1 & 14 & 0 & 0 \\
    train & 18 & 16 & 42 & 13 & 4 & 7 & 0 & 0 \\
    trout & 20 & 6 & 15 & 28 & 14 & 16 & 1 & 0 \\
    tulip & 53 & 2 & 11 & 4 & 16 & 13 & 1 & 0 \\
    turtle & 14 & 9 & 37 & 10 & 6 & 22 & 2 & 0 \\
    wardrobe & 10 & 22 & 10 & 20 & 9 & 20 & 2 & 7 \\
    whale & 1 & 4 & 85 & 7 & 0 & 3 & 0 & 0 \\
    willow_tree & 13 & 3 & 43 & 26 & 2 & 5 & 0 & 8 \\
    wolf & 17 & 10 & 22 & 5 & 22 & 18 & 6 & 0 \\
    woman & 12 & 21 & 16 & 19 & 7 & 15 & 8 & 2 \\
    worm & 13 & 13 & 26 & 8 & 17 & 15 & 7 & 1 \\
    \bottomrule
  \end{tabular}
  \end{adjustbox}
  \caption{\textbf{CIFAR100-B Statistics (part 2).} The number of instances that belongs to each background color from each class.}
  \label{tab:cifar100b_appendix}
\end{table*}

\end{document}